\documentclass[12pt]{iopart}
\usepackage[utf8]{inputenc}
\usepackage[T1]{fontenc}
\usepackage{graphicx}
\usepackage{subcaption}
\usepackage{multirow}
\usepackage{hhline}
\usepackage{ctable}
\usepackage{algpseudocode}
\usepackage{algorithm}
\usepackage{url}
\usepackage{hyperref}
\usepackage{harvard}
\usepackage{fixltx2e}

\MakeRobust{\Call}
\algnewcommand\algorithmicinput{\textbf{INPUT:}}
\algnewcommand\INPUT{\item[\algorithmicinput]}

\algnewcommand\algorithmicoutput{\textbf{OUTPUT:}}
\algnewcommand\OUTPUT{\item[\algorithmicoutput]}

\hypersetup{colorlinks=false,pdfborder={0 0 0}, bookmarksopen,bookmarksdepth=3}

\begin{document}


\title[Abductive reasoning for Atrial Fibrillation identification]{Abductive 
reasoning as the basis to reproduce expert criteria in ECG Atrial Fibrillation 
identification.}

\author{T Teijeiro, C A García, D Castro and P Félix}

\address{Centro Singular de Investigación en Tecnoloxías da Información 
(CiTIUS), University of Santiago de Compostela, Spain}
\ead{tomas.teijeiro@usc.es}

\begin{abstract}
\textit{Objective:} This work aims at providing a new method for the automatic 
detection of atrial fibrillation, other arrhythmia and noise on short single 
lead ECG signals, emphasizing the importance of the interpretability of the 
classification results. \textit{Approach:} A morphological and rhythm 
description of the cardiac behavior is obtained by a knowledge-based 
interpretation of the signal using the \textit{Construe} abductive framework. 
Then, a set of meaningful features are extracted for each individual heartbeat 
and as a summary of the full record. The feature distributions were used to 
elucidate the expert criteria underlying the labeling of the 2017 Physionet/CinC 
Challenge dataset, enabling a manual partial relabeling to improve the 
consistency of the classification rules. Finally, state-of-the-art machine 
learning methods are combined to provide an answer on the basis of the feature 
values. \textit{Main results:} The proposal tied for the first place in the 
official stage of the Challenge, with a combined $F_1$ score of 0.83, and was
even improved in the follow-up stage to 0.85 with a significant simplification 
of the model. \textit{Significance:} This approach demonstrates the potential of 
\textit{Construe} to provide robust and valuable descriptions of temporal data 
even with significant amounts of noise and artifacts. Also, we discuss the 
importance of a consistent classification criteria in manually labeled training 
datasets, and the fundamental advantages of knowledge-based approaches to 
formalize and validate that criteria.
\end{abstract}



\submitto{\PM}

\maketitle

\section{Introduction}

In the last decades, the capacity of Artificial Intelligence to provide low-cost 
methods for the automatic diagnosis of cardiac diseases using standard ECG 
records has been a recurrent claim. There are thousands of works from many 
different approaches addressing various common problems, such as feature 
extraction, beat classification or rhythm analysis, among 
others~\cite{Sornmo05}. Nevertheless, the potential shown in research results is 
still largely untapped in the clinical routine, and most of the current analysis 
tasks require an intensive intervention of expert clinicians. The 
Physionet/Computing in Cardiology Challenge tries every year to reduce this gap 
for some recognized problem, and in the 2017 edition it defied the scientific 
community and the industry to propose viable solutions to provide a reliable 
screening of Atrial Fibrillation from short single-lead ECG signals acquired 
with a commercial low-cost device~\cite{Clifford17}. A total of 8528 ECG records 
were provided as a training set, labeled in four classes: Normal sinus rhythm 
(\textbf{N}), Atrial fibrillation (\textbf{A}), Other rhythm (\textbf{O}) and 
Noisy ($\sim$). A hidden test of 3658 records were used to evaluate the 
performance of the proposed algorithms, using as metric the mean $F_1$ measure 
of the \textbf{N}, \textbf{A}, and \textbf{O} classes.

Besides pursuing a competitive numeric accuracy, the present work makes special 
emphasis on the interpretability of the results, as this has proven to be a 
major concern of care staff to trust automatic assistance 
methods~\cite{Caruana15}. For this, the classification procedure is based on a 
set of high-level features obtained from the description of the ECG in the same 
terms used by expert clinicians. This description is generated by the 
\textit{Construe} algorithm~\cite{Teijeiro16}, which relies on abductive 
reasoning to obtain the best interpretation of the observed evidence, using a 
knowledge-based approach.

One of the main difficulties recognized by the participants in this Challenge 
is the absence of specific classification criteria beyond the name of each 
class. Even the expert clinicians labeling the training and test datasets 
received no instructions, leading to a large level of disagreement and many 
label inconsistencies~\cite{Clifford17}. Thus, machine learning methods hit a 
performance barrier that was below what it could be expected a priori. In our 
case, we tried to overcome this limit by elucidating the expert criteria 
underlying the training dataset and improving the label consistency according to 
that criteria, rather than tuning the learning algorithms. This strategy could 
even be improved in the follow-up stage of the Challenge, after a significant 
change of criteria between the training and test sets was unveiled in the data 
profile~\cite{Clifford17}.

\begin{figure*}[t]
\centering
\includegraphics[width=\textwidth]{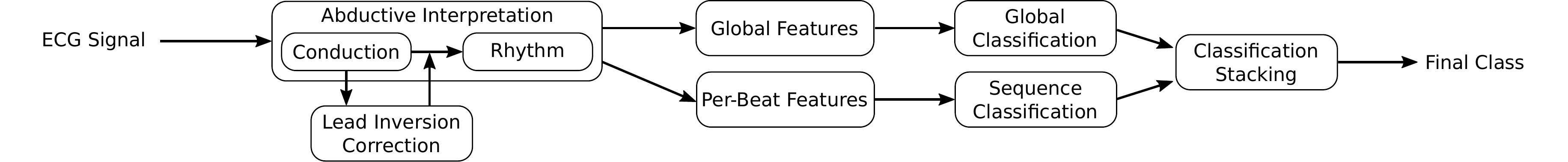}
\caption{Algorithm architecture}
\label{fig:arch}
\end{figure*}

The proposed classification algorithm is a combination of an abductive 
knowledge-based approach to interpret the raw signal with some learning-based 
methods that fit the decision parameters to the Challenge training set. 
Figure~\ref{fig:arch} shows the global architecture of the algorithm, that is 
described in detail in section~\ref{sec:methods} along with the manual data 
relabeling process. Afterwards, section~\ref{sec:results} shows the validation 
results obtained on the Challenge training set and in the official and follow-up 
stages. Finally, section~\ref{sec:conclusions} discusses these results, the 
potential of the proposed approach and a possible roadmap for the development of 
a low-cost automatic screening method for atrial fibrillation and other possible 
target arrhythmias.

\section{Methods}
\label{sec:methods}

\subsection{Abductive interpretation of the ECG}
\label{subsec:abduction}

The abductive interpretation of the ECG is the most distinguishing feature of 
our proposal. The result of this interpretation is a description in multiple 
abstraction levels of the physiological processes underlying the signal trace, 
in the same terms used by expert clinicians. Two abstraction levels are 
considered: the \textit{conduction} level, that describes the signal as a 
sequence of P waves, QRS complexes and T waves, each with its corresponding 
morphology, amplitude and duration; and the \textit{rhythm} level, that 
describes the signal as a concatenation of segments showing different rhythm 
patterns, including normal sinus rhythm, bradycardia, tachycardia, extrasystole, 
couplet, rhythm block, bigeminy, trigeminy, atrial fibrillation, ventricular 
flutter and asystole. 

To build this description, the \textit{Construe} algorithm implements a 
hypothesize-and-test cycle that pursues the best explanation of the observed 
evidence according to the available knowledge. The domain knowledge is defined 
as a set of \textit{abstraction patterns} that describe the time and value 
constraints that have to be satisfied by evidence \textit{observables} to 
support a given hypothesis. Abstraction patterns are dynamically generated from 
formal descriptions named \textit{abstraction grammars}. The knowledge base for 
the interpretation of ECG signals is described in~\cite{Teijeiro16}. 

The results of the abductive interpretation are also used to detect and fix 
possible inversions in the ECG signal. Lead inversion was found to be a quite 
common issue in the training set, affecting approximately 15\% of the records. 
This is probably due to an incorrect holding of the acquisition device, and as a 
consequence it increases the chances to classify a record as abnormal due to the 
presence of infrequent QRS and T wave morphologies, as well as to the greater 
difficulty to identify the P waves.

This detection is performed from the \textit{Construe} results at the conduction 
level and before carrying out the rhythm interpretation. The initial evidence 
are the beat annotations produced by the \texttt{gqrs} application from the 
Physionet library~\cite{Goldberger00}, from which a tentative delineation of the 
P wave, QRS complex and T wave of each heartbeat is obtained.

The inverted records were first identified manually, and then a simple logistic 
regression classifier was trained considering 14 features obtained from the raw 
signal and the delineation results. These features are 1) the median of the QRS 
axis; 2) the median of the QRS amplitude; 3) the difference between the mean and 
the median of the signal, normalized by the signal length and 4) by the signal 
amplitude; 5) the baseline value, calculated as the mode of the signal; 6) the 
ratio of the energy of the signal above and below the baseline; 7) the 
dispersion of the signal; 8) the mean value of the signal samples exceeding more 
than 3 standard deviations from the baseline; 9) the median amplitude of the 
first, 10) second, and 11) third waves inside the QRS complex; and the number of 
12) P waves, 13) QRS complexes, and 14) T waves normalized by the signal length. 
The classifier showed a cross-validation F1 score of 0.96. If the signal is 
found to be inverted, all signal samples are multiplied by -1 before continuing.

\begin{figure}[t]
\centering
\includegraphics[width=\textwidth]{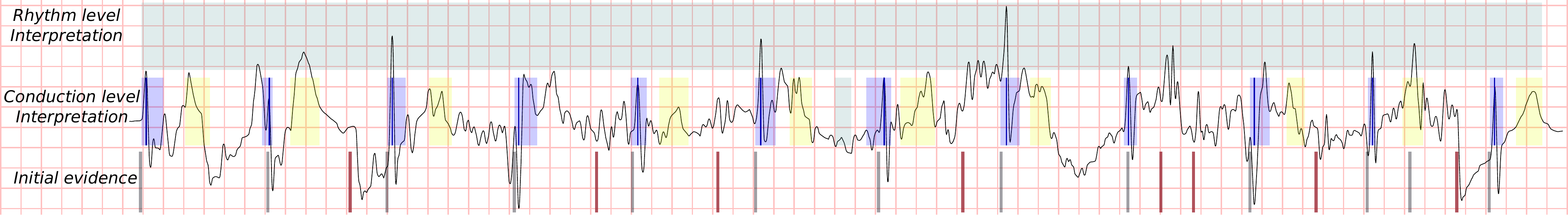}
\caption{Fixing false positive detections. {\scriptsize [Source: First 10 
seconds of the A02080 record. Grey: Explained \texttt{gqrs} annotations. Red: 
Discarded \texttt{gqrs} annotations. Blue: QRS observations. Yellow: T wave 
observations. Green: P wave observation and Normal rhythm hypothesis.]}} 
\label{fig:fp_fix}
\bigskip
\centering
\includegraphics[width=\textwidth]{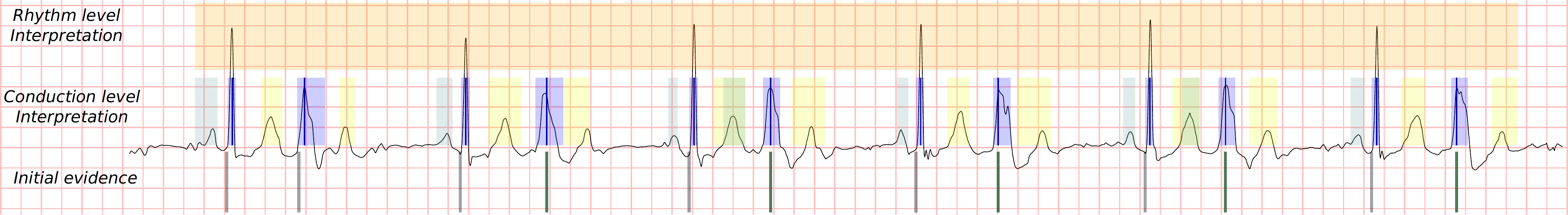}
\caption{Fixing false negative detections. {\scriptsize [Source: Record 
A01744 from second 4 to 14. Grey: Original \texttt{gqrs} annotations. Blue: QRS 
observations. Yellow: T wave observations and Bigeminy hypothesis. Green: 
Discovered beat annotations and P wave observations.]}} \label{fig:fn_fix}
\end{figure}

After the lead inversion correction step, the \textit{Construe} algorithm is 
executed up to the rhythm level taking as initial evidence the same set of QRS 
complexes used in the delineation step. This set can be modified during the 
interpretation process thanks to the non-monotonic reasoning scheme that 
combines bottom-up reasoning (guided by data) and top-down reasoning (guided by 
knowledge) to obtain the best matching between observations and knowledge. This 
gives great robustness to the presence of noise and artifacts in the ECG data, 
and allows to fix both false positive and false negative QRS detections, as 
shown in Figures~\ref{fig:fp_fix} and~\ref{fig:fn_fix}. In 
Figure~\ref{fig:fp_fix}, the high amount of noise in the signal causes many 
false positive beat detections, that are fixed at the rhythm level by selecting 
as the best explaining hypothesis a single normal rhythm observation. In 
Figure~\ref{fig:fn_fix}, all but the first ventricular beat are not detected, 
probably due to the lower amplitude and slope. Still, the bigeminy hypothesis 
evoked by the first extrasystole allows to look for further advanced beats, 
explaining the full fragment with this rhythm hypothesis.

\subsection{High level feature extraction}

The description resulting from the abductive interpretation stage contains the 
information typically referred by expert clinicians and ECG handbooks to make a 
decision about the normal/abnormal condition of a record. However, since the 
criteria leading to each label is not provided, this information is converted to 
a set of quantitative features that will be the input for two selected machine 
learning methods in order to fit the underlying criteria of the training set. 
One of the methods evaluates each record globally, using aggregated features; 
while the other evaluates the record as a sequence, using a Recurrent Neural 
Network architecture fed with individual features for each detected heartbeat. 
These two methods are detailed below.

In the proposal submitted to the official phase of the 
Challenge~\cite{Teijeiro17}, a total of 79 features were calculated. Yet, the 
final results obtained in the test set showed a significant overfitting of the 
training set, which is a sign of an excessive complexity of the model. Thus, in 
the new version the set of global features was reduced using the Recursive 
Feature Elimination technique~\cite{Guyon02}, resulting in 42 features that are 
detailed in Table~\ref{tab:features}. Two of these features, \texttt{RR\_Irr} 
and \texttt{RR\_MIrr}, were not included in the previous version and correspond 
to a rhythm irregularity measure based on sample entropy estimation, which is 
described in~\cite{Petrenas15}. 

Some of these features require further explanation. Specifically, the term 
\textit{profile} refers to the sum of the absolute value of the derivative 
signal, and is an excellent signal quality indicator. Regarding \texttt{Pdistd} 
and \texttt{MPdist}, the referred measure corresponds to the distance to the 
separating hyperplane of the one class SVM used by \textit{Construe} to perform 
the delineation of P waves, and intuitively it is a measure of ``how much it 
really looks like a P wave''.

Most of the individual features used to train the sequence classifier described 
in section~\ref{subsec:sequence_classification} are just disaggregations of the 
features in Table~\ref{tab:features}, such as \texttt{RR, RR\_Irr, n\_PR, prof, 
pw\_prof, Pdist, QT, TP} and \texttt{TP\_freq}. Additionally, various 
morphological features were added to enable the identification of isolated 
conduction alterations. These features include the duration, amplitude and 
turning point of each individual wave inside the QRS complex, the QRS axis, the 
P and T waves duration and amplitude, and the ST segment deviation. Also, two 
qualitative features were included to describe the QRS morphology tag 
(\textit{qRs, QS, rSr', etc.})~\cite{Teijeiro15} and the name of the rhythm in 
which the heartbeat was interpreted.

\begin{table}
\renewcommand{\arraystretch}{1.2}
\caption{Set of features used to train the global classifier}
\label{tab:features}
\scriptsize
\begin{center}
\begin{tabularx}{\textwidth}{|X|X|}
\hline
\texttt{tSR:} Proportion of the record length interpreted as a regular 
rhythm (Normal rhythm, tachycardia or bradycardia). & \texttt{t1b:} Number of 
milliseconds from the beginning of the record to the first interpreted 
heartbeat. \\
\hline
\texttt{tOR:} Number of milliseconds interpreted as a non-regular rhythm. & 
\texttt{longTch:} Longest period of time with heart rate over 100bpm. \\
\hline
\texttt{RR:} Median RR interval of regular rhythms. & 
\texttt{RRd\_std:} Standard deviation of the instant RR variation.\\
\hline
\texttt{RRd:} Median Absolute Deviation (MAD) of the RR interval in regular 
rhythms. &
\texttt{MRRd:} Max. absolute variation of the RR interval in regular rhythms.\\
\hline
\texttt{RR\_MIrr:} Max. RR irregularity measure. & \texttt{RR\_Irr:} Median RR 
irregularity measure.\\
\hline
\texttt{PNN\{10,50,100\}:} Global PNNx measures. & \texttt{o\_PNN50:} PNN50 of 
non-regular rhythms.\\
\hline
\texttt{mRR:} Min. RR interval of regular rhythms. & 
\texttt{o\_mRR:} Min. RR interval of non-regular rhythms.\\
\hline
\texttt{n\_nP:} Proportion of heartbeats with detected P-wave inside regular 
rhythms. & \texttt{n\_aT:} Median of the amplitude of the T waves inside 
regular rhythms.\\
\hline
\texttt{n\_PR:} Median PR duration inside regular rhythms. & \texttt{Psmooth:} 
Median of the ratio between the standard deviation and the mean value of 
P-waves' derivative signal.\\
\hline
\texttt{Pdistd:} MAD of the measure given by the P wave delineation method. & 
\texttt{MPdist:} Max. of the measure given by the P wave delineation method.\\
\hline
\texttt{prof:} Profile of the full signal. & \texttt{pw\_profd:} MAD of 
\texttt{pw\_prof}.\\
\hline
\texttt{xcorr:} Median of the maximum cross-correlation between QRS complexes 
interpreted in regular rhythms. & \texttt{o\_xcorr:} Median of the maximum 
cross-correlation between QRS complexes interpreted in non-regular rhythms.\\
\hline
\texttt{PRd:} Global MAD of the PR durations. & \texttt{QT:} Median of the 
corrected QT measure.\\
\hline 
\texttt{TP:} Median of the prevailing frequency in the TP intervals. &
\texttt{TPfreq:} Median of the frequency entropy in the TP intervals.\\
\hline
\texttt{pw\_prof:} Profile measure of the signal in the P-wave area. & 
\texttt{nT:} Proportion of QRS complexes with detected T waves.
\\
\hline
\texttt{n\_Txcorr:} Median of the maximum cross-correlation between T-waves 
inside regular rhythms. & \texttt{n\_Pxcorr:} Median of the maximum 
cross-correlation between P-waves inside regular rhythms.\\
\hline
\texttt{baseline:} Profile of the baseline in regular rhythms. & 
\texttt{o\_baseline:} Profile of the baseline in non-regular rhythms.\\
\hline
\texttt{wQRS:} Proportion of wide QRS complexes (duration longer than 110ms). & 
\texttt{wQRS\_xc:} Median of the maximum cross-correlation between wide QRS 
complexes.\\
\hline
\texttt{wQRS\_prof:} Median of the signal profile in the 300ms before each wide 
QRS complex. & \texttt{w\_PR:} Proportion of heartbeats with long PR interval 
(longer than 210 ms).\\
\hline
\texttt{x\_xc:} Median of the maximum cross-correlation between ectopic beats. 
& \texttt{x\_rrel:} Median of the ratio between the previous and next RR 
intervals for each ectopic beat.\\
\hline
\end{tabularx}
\end{center}
\vspace{-1.5em}
\end{table}

\subsection{Expert criteria elucidation and data relabeling}

Every non-random human labeling task has a set of underlying criteria. These 
criteria may be subtle, unconscious, or even variable along time, but they 
necessary exist. The objective of any machine-learning algorithm trying to 
reproduce this labeling will therefore be to formalize and quantify these 
underlying criteria, and in consequence, the more formalized these criteria are 
a priori, the easier will be the development of such an algorithm. Probably, 
the most labor-intensive task of our proposal was the elucidation of the expert 
criteria underlying the training and test sets, and the ensuing data relabeling 
to make these criteria as consistent as possible along the dataset.

Certainly, the most difficult class to define an appropriate discrimination 
knowledge is the \textbf{O} class, inasmuch as the only provided information 
(the class name) is excessively vague and it may include a range of 
pathophysiological processes showing very different morphologies and rhythms. 
Hence, since this class is opposed to atrial fibrillation and normal sinus 
rhythm, one expert may consider that only rhythm alterations should be included 
in this class, while another expert may contemplate any event that is out of 
normality, such as conduction delays or chamber enlargements, among others. 

\begin{figure*}
  \centering
  \includegraphics[width=0.3\textwidth]{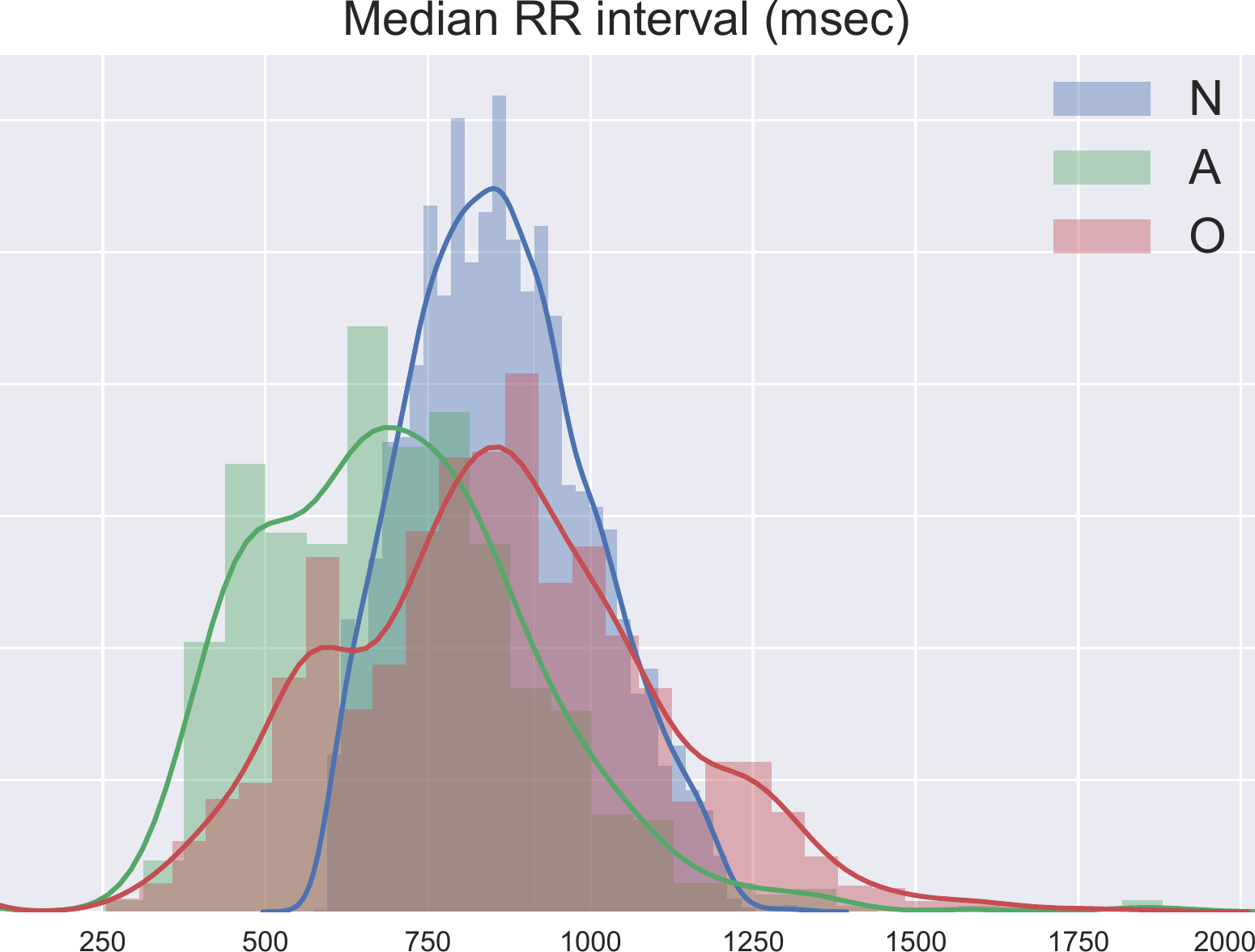}\hspace{1.5em}
  \includegraphics[width=0.3\textwidth]{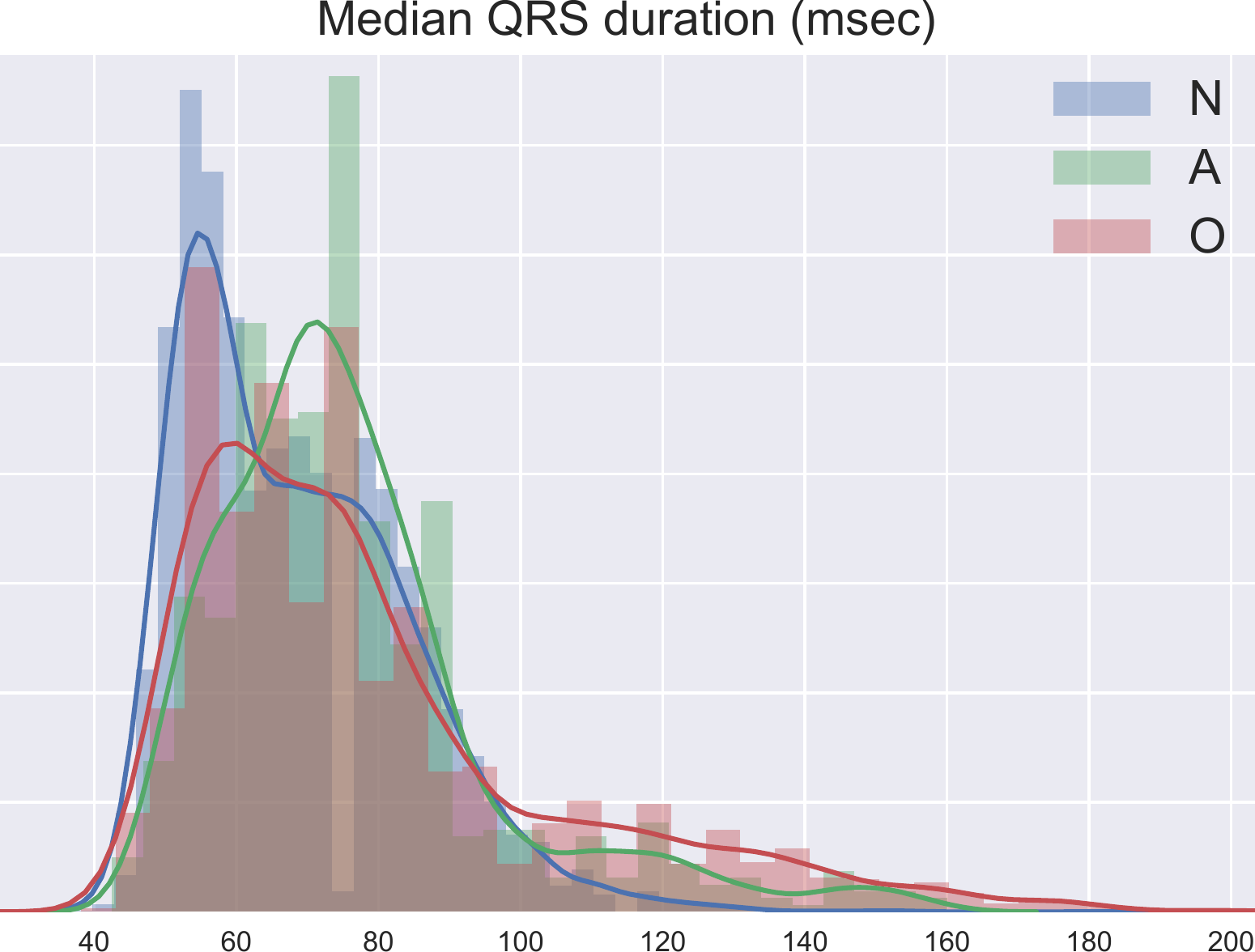}\hspace{1.5em}
  \includegraphics[width=0.3\textwidth]{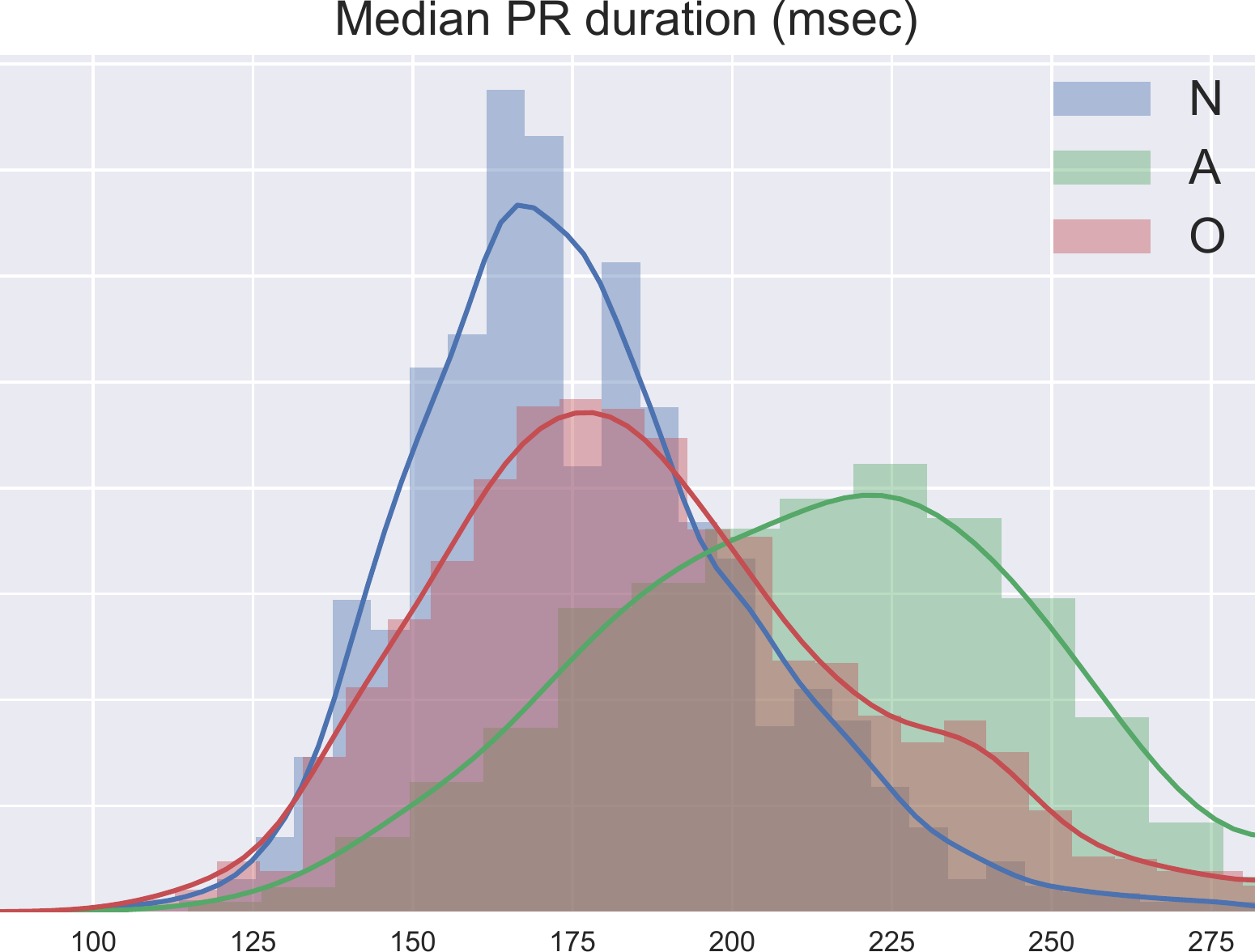}
  \caption{Per-class distribution of three representative features (best seen 
in color).}
  \label{fig:kde}
  \vspace{-1.5em}
\end{figure*}

Thanks to the physiological meaning of the features provided by the 
interpretation, it has been possible to throw light on some well-known ECG 
alterations that seem to be considered as \textbf{O} representatives in the 
training set. A simple but valuable tool is the per-class distribution of each 
feature. Figure~\ref{fig:kde} shows the distributions of three  features that 
define an almost crisp threshold for some well-known anomalies. It can be seen 
that the tachycardia and bradycardia abnormal rhythms are considered to have a 
median RR interval below 600ms and over 1200ms, approximately. Also, a QRS 
complex wider than 110ms seems a clear indicator of anomaly, and a PR duration 
over 220ms also has a bias to the \textbf{O} class with respect to \textbf{N}, 
although in this case the separation is less clear-cut.

Using this procedure, the following recognized rhythm and conduction conditions 
were identified to be labeled as \textbf{O} in the training set (a 
representative record of each anomaly is also displayed, corresponding to 
\textbf{A07833, A05308, A06071, A00688, A05301, A06295, A00741} and 
\textbf{A00326} records):

\begin{center}
\small
\renewcommand{\arraystretch}{1.5}
\setlength{\tabcolsep}{0.35em}
\begin{tabular}{m{0.27\textwidth} m{0.73\textwidth}}
 $\bullet$ Tachycardia & \includegraphics[width=0.7\textwidth]{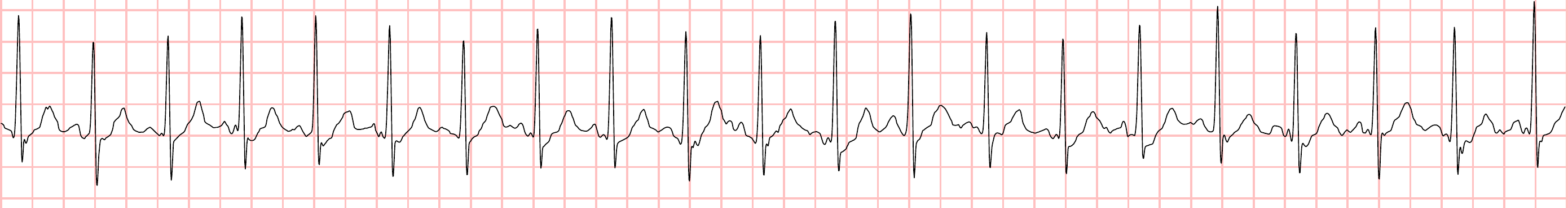}\\
 $\bullet$ Bradycardia & \includegraphics[width=0.7\textwidth]{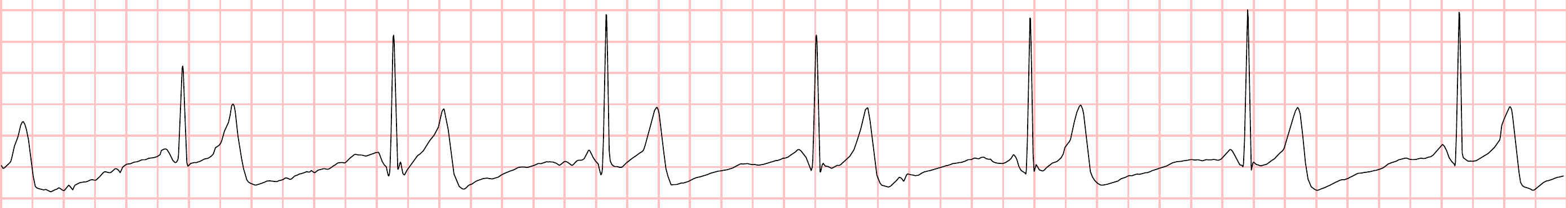}\\
 $\bullet$ Wide QRS complex & \includegraphics[width=0.7\textwidth]{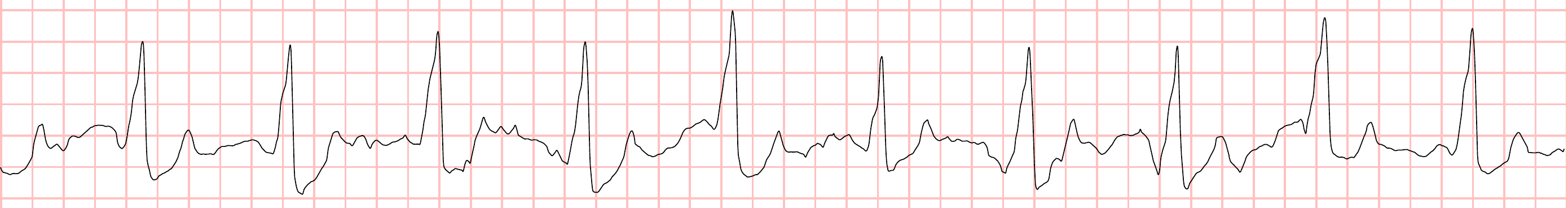}\\
 $\bullet$ Presence of ventricular or fusion beats & 
\includegraphics[width=0.7\textwidth]{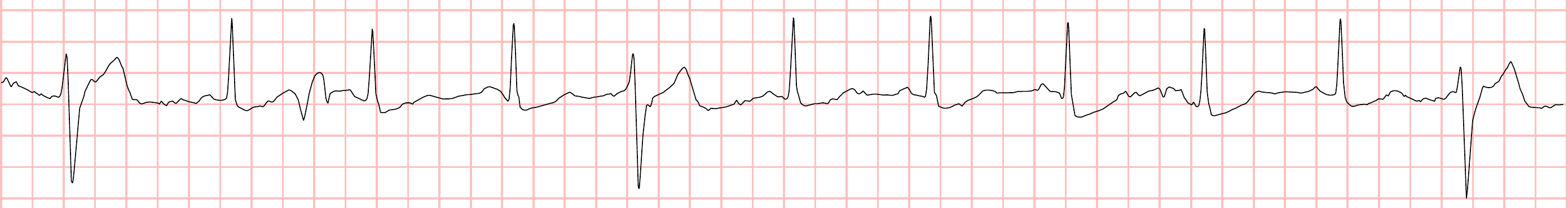}\\
 $\bullet$ Presence of at least one extrasystole  & 
\includegraphics[width=0.7\textwidth]{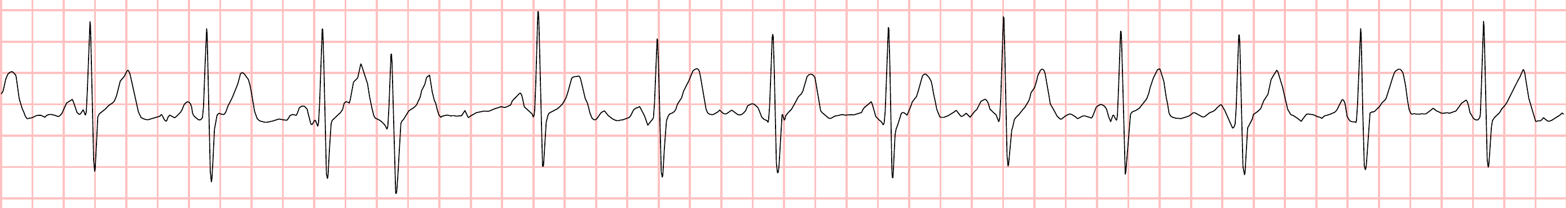}\\
 $\bullet$ Long PR interval & \includegraphics[width=0.7\textwidth]{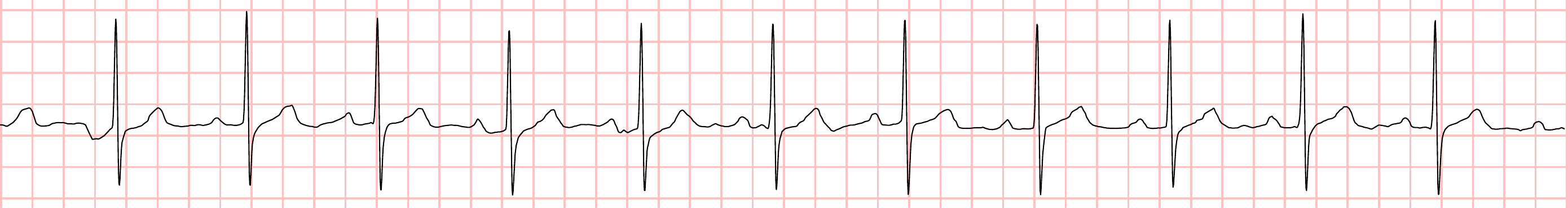}\\
 $\bullet$ Ventricular tachycardia & 
\includegraphics[width=0.7\textwidth]{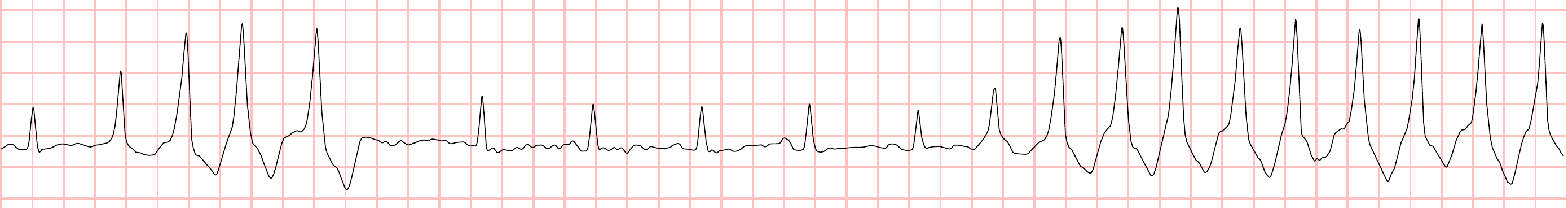}\\
 $\bullet$ Atrial flutter  & \includegraphics[width=0.7\textwidth]{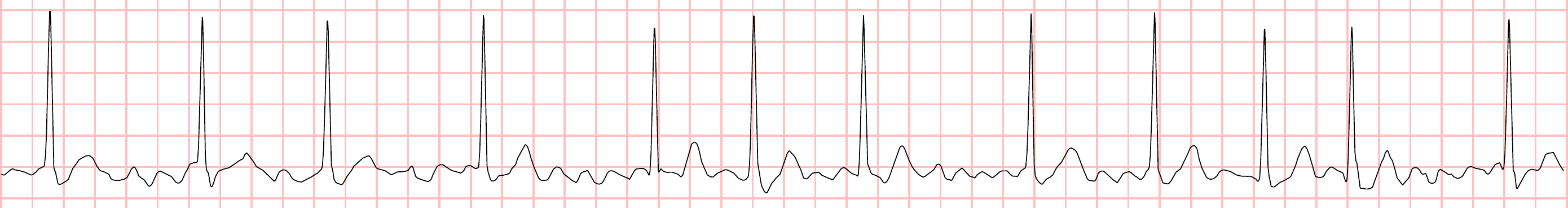}\\
\end{tabular}
\end{center}

\begin{figure}[h!!]
 \centering
 \begin{tabular}{cc}
  \begin{subfigure}{0.45\textwidth}
    \includegraphics[width=\textwidth]{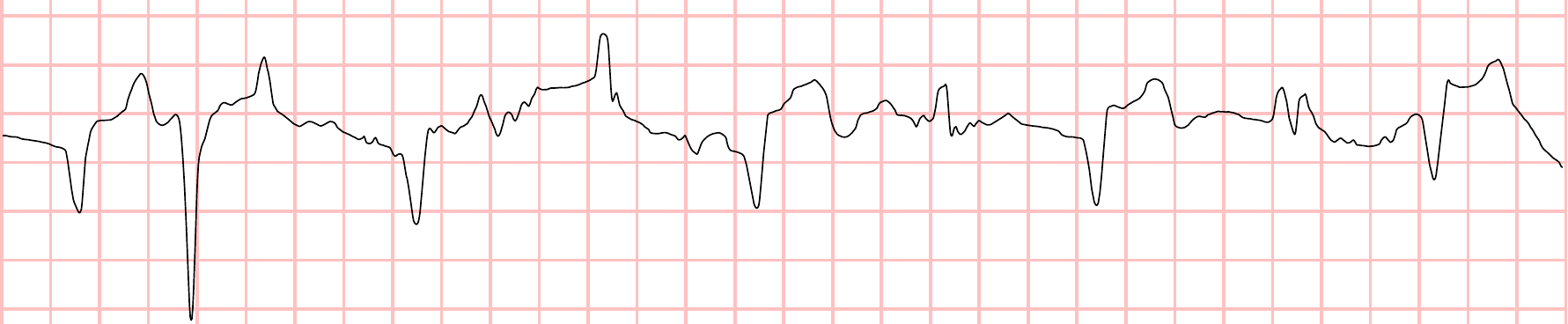}
    \caption{\scriptsize{\textbf{A06681:} Ventricular beats labeled as 
\textbf{N}}}
  \end{subfigure}&
  \begin{subfigure}{0.45\textwidth}
    \includegraphics[width=\textwidth]{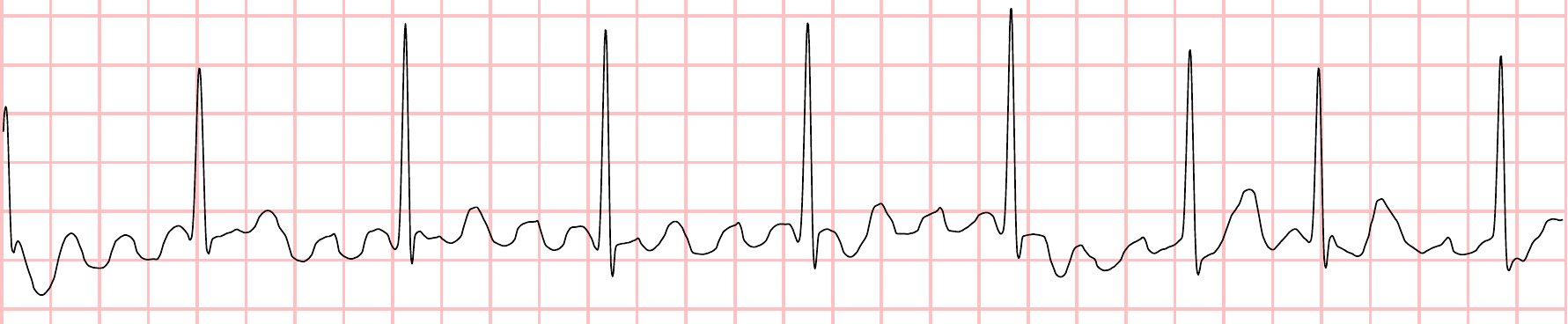}
    \caption{\scriptsize{\textbf{A07836:} Atrial flutter labeled as \textbf{A}}}
    \label{subfig:flutter}
  \end{subfigure}\vspace{0.5em}\\
 
  \begin{subfigure}{0.45\textwidth}
    \includegraphics[width=\textwidth]{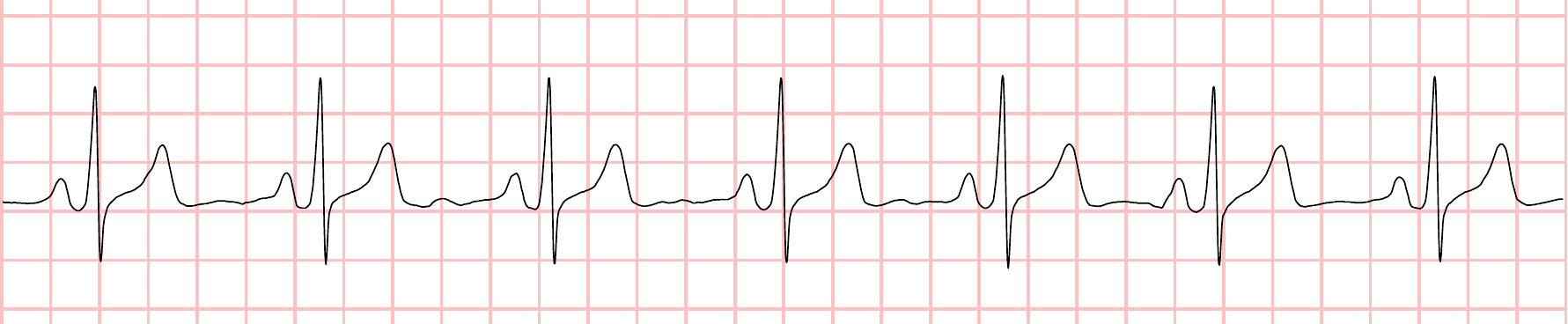}
    \caption{\scriptsize{\textbf{A00848:} Normal record labeled as \textbf{O}}}
  \end{subfigure} &
  \begin{subfigure}{0.45\textwidth}
    \includegraphics[width=\textwidth]{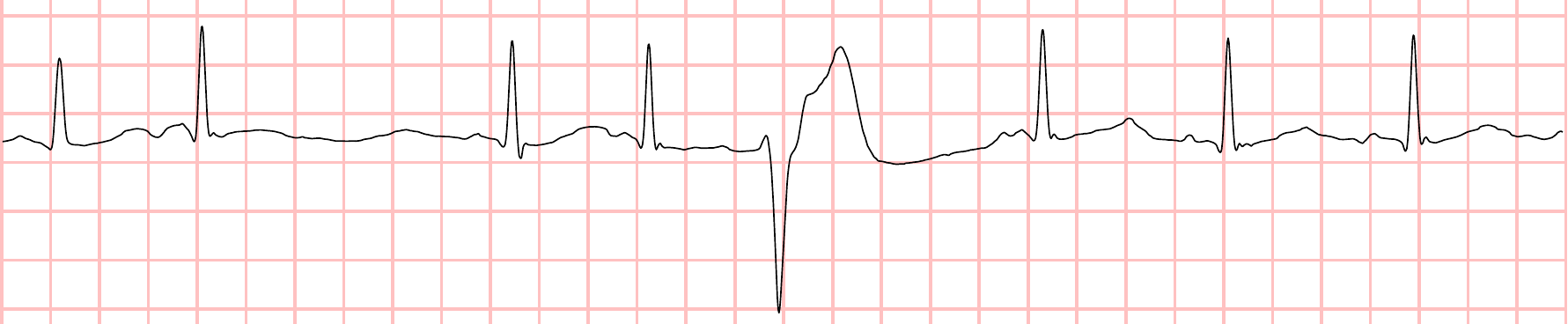}
    \caption{\scriptsize{\textbf{A03934:} Extrasystoles labeled as $\sim$}\label{fig:incosistencies_d}}
  \end{subfigure}
  \end{tabular}
 \caption{Inconsistent labels in the training set}
 \label{fig:inconsistencies}
\end{figure}

Each of these conditions is sufficient to classify a record as \textbf{O} in 
many cases, such as those shown above. However, these criteria are not 
consistent across the entire dataset and there are examples of these phenomena 
in several classes. Some examples are shown in Figure~\ref{fig:inconsistencies}.

Inconsistencies seriously undermine the performance of machine learning 
algorithms, inasmuch as trying to fit arbitrary decisions lead to more complex 
and overfitted models. Hence, a partial manual relabeling process guided by 
misclassified examples was performed for the official phase of the Challenge. 
This process tried to be conservative and focused on clear examples of the 
abovementioned anomalies that were not labeled as \textbf{O}. A total number of 
197 out of 8528 records were relabeled in this stage, named V2.

During the official phase, a significant portion of the records in the hidden 
test set was relabeled by up to eight human experts, following a bootstrap 
approach based on the disagreement level of the top scored algorithms. Some 
basic statistics about this process were published by the organizers~\cite[Table 
2]{Clifford17}, showing an important reduction of the records labeled as 
\textbf{O}, with a relative decrease of about 33\%. This suggest a significant 
change of the underlying classification criteria, being more conservative 
towards the \textbf{N} and $\sim$ classes. Since labels in the training set were 
not modified accordingly, cross-validation was no longer a reliable predictor of 
the performance of the algorithms in the test set, and intuition became much 
more relevant.

Following this intuition, a second relabeling stage was performed for the 
follow-up phase of the Challenge. It is reasonable to think that the change of 
criteria in favor of the Normal and Noise classes had a greater impact on 
records showing a regular sinus rhythm, so the conduction criteria (wide QRS 
complex and long PR interval) were no longer considered sufficient for the 
\textbf{O} class. Also, and taking into account that the signal was captured by 
subjects participating actively and with a handheld device, phenomena like 
ventricular tachycardias were considered implausible, and patterns like those 
appearing in record A00741 were labeled as noise. A total number of 747 records 
were relabeled in this stage, named V3. Table~\ref{tab:class_stats} shows a 
comparison of the class distributions in the training set before and after 
relabeling for each phase of the challenge, and the distribution in the hidden 
test set.

\begin{table}[h!]
\caption{Class distribution before/after manual relabeling}
\label{tab:class_stats}
\begin{center}
\footnotesize
\begin{tabular}{|c|r|r|r|r|r|}\hline
Class  & Official V2 & Relabeled V2 & Official V3 & Relabeled V3 & Test 
Set\\\hline
\textbf{N} & 59.2\% & 57.4\% & 59.5\% & 64.8\% & 66.6\%\\\hline
\textbf{A} & 8.7\% & 8.6\% & 8.9\% & 8.6\% & 7.8\%\\\hline
\textbf{O} & 28.8\% & 30.7\% & 28.3\% & 23.0\% & 18.7\%\\\hline
$\sim$ & 3.3\% & 3.2\% & 3.3\% & 3.6\% & 6.9\%\\\hline
\end{tabular}
\end{center}
\vspace{-1.5em}
\end{table}

Additionally, the heuristics guiding the \textit{Construe} algorithm were 
slightly modified to encourage normal rhythm hypotheses. Specifically, if a 
signal fragment shows evidence compatible with the presence of QRS complexes at 
regular intervals and with a frequency between 50 and 110 beats per minute, 
then the fragment is considered fully explained and no other rhythm hypotheses 
are explored. This adjustment can potentially ignore important anomalies such 
as interpolated extrasystoles, but it was observed that with a single lead it 
is infeasible to morphologically distinguish many artifact patterns from real 
ectopic QRS complexes, so the simplicity principle is enforced in this case.

\subsection{Global classification}

As shown in Figure~\ref{fig:arch}, the label for a given record is obtained by 
combining two different classification algorithms. The first one is a ``global'' 
classifier, that makes a decision based on a summary information of the full 
record. It is defined as a standard multilabel classifier $\mathcal{F: X 
\rightarrow Y}$, where $\mathcal{X}$ is the feature space defined in 
Table~\ref{tab:features} and $\mathcal{Y} = \{\mathbf{N, A, O, \sim}\}$.

The selected machine learning method was the Tree Gradient Boosting algorithm, 
and particularly the XGBoost implementation~\cite{Chen16}, which achieved 
significantly better results than other state-of-the art methods such as Support 
Vector Machines or Random Forests without any hyperparameter tuning. Also, this 
algorithm provides a certain degree of interpretability by assigning different 
importances to the classification features. Table~\ref{tab:alg_comparison} shows 
a basic comparison of the $F_1$ score obtained for each class and for the final 
metric by each of these algorithms on the training set after relabeling, using 
200-fold cross-validation. For SVM and Random Forest, the scikit-learn v0.18
implementation was used~\cite{sklearn}. 

\vspace{-1em}

\begin{table}[h!!]
\caption{Algorithm comparison for global classification without hyperparameter 
tuning}
\label{tab:alg_comparison}
\footnotesize
\begin{center}
\begin{tabular}{|c|r|r|r|}\hline
           & SVM  & Random Forest & XGBoost\\\hline
\textbf{N} & 0.95 &          0.95 &    0.96\\\hline
\textbf{A} & 0.80 &          0.77 &    0.81\\\hline
\textbf{O} & 0.81 &          0.82 &    0.86\\\hline
$\sim$     & 0.67 &          0.63 &    0.69\\\hline
\textbf{Overall} & \textbf{0.85} & \textbf{0.85} & \textbf{0.88}\\\hline
\end{tabular}
\end{center}
\end{table}

%
%
%

%
%
%

%
%
%

Hyperparameter tuning for XGBoost was performed using exhaustive grid search and 
8-fold cross-validation. The tested and selected values are shown in 
Table~\ref{tab:hyperparameters}.

\vspace{-1em}

\begin{table}[h!]
\caption{Hyperparameters tested for XGBoost optimization}
\label{tab:hyperparameters}
\begin{center}
\footnotesize
\renewcommand{\arraystretch}{1.2}
\begin{tabular}{|l|l|}
\hline
Maximum tree depth: \textbf{5}, 6, 7, 10 and 12. & 
Gamma: 0.2, 0.4, 1.0 and \textbf{2.0}.\\
\hline
Learning rate: \textbf{0.1}, 0.2 and 0.4. & 
Subsample: \textbf{0.8}, 0.9 and 1.0.\\
\hline
Number of boosting rounds: 25, 50, 75 and \textbf{100}. &
 Min. child weight: 5, \textbf{7}, 10 and 15.\\
\hline
\end{tabular}
\end{center}
\end{table}

\vspace{-1em}

The obtained results gave a best mean $F_1$ score of 0.886 $\pm$ 0.015. The 
worst result was 0.858 $\pm$ 0.008, but more than 90\% of the hyperparameter 
combinations were within the standard deviation of the best result, that is, 
over 0.871. As the variability is so low, we decided to choose conservative 
values for most of the hyperparameters to build a more general model. The 
selected values are highlighted in bold, and gave a score of 0.883 $\pm$ 0.010. 
On the training set, the global classifier achieved a score of 0.942.

\subsection{Sequence classification}
\label{subsec:sequence_classification}
The second classification algorithm shown in Figure~\ref{fig:arch} was conceived 
as a complement to the global classifier, by working in a beat-by-beat basis 
instead of dealing with features summarizing the whole record. Consider, for 
example, an almost normal record with just an ectopic beat. Unless the summary 
of the record and the global classifier are extremely accurate, these type of 
records will probably be misclassified. However, a classifier specialized in 
processing sequences of beats could correctly classify the record by remembering 
that an abnormal heartbeat has occurred. There is another compelling reason for 
using a sequence classifier. Although \textit{Construe} is able to extract 
meaningful features in noisy segments, it is reasonable to give more credence to 
clean measurements than to noisy ones. Unlike with global features, this is easy 
to achieve with the sequence classifier, in which each heartbeat receives its 
own signal quality indicator.

The need for remembering the occurrence of (probable) distant events led us to 
use Long Short Term Memory networks (LSTMs) as the basis of the sequence 
classifier. The key idea behind LSTMs is the use of a cell memory which can 
store useful information over time, and several non-linear gating units that 
decide which information should be added and removed from the cell. The use of 
the gating cells is also important to avoid vanishing or exploding gradients, an 
important issue when backpropagating through time and which may prevent the 
network from learning long-range dependencies.

\begin{figure*}[t]
\centering
\includegraphics[width=\textwidth]{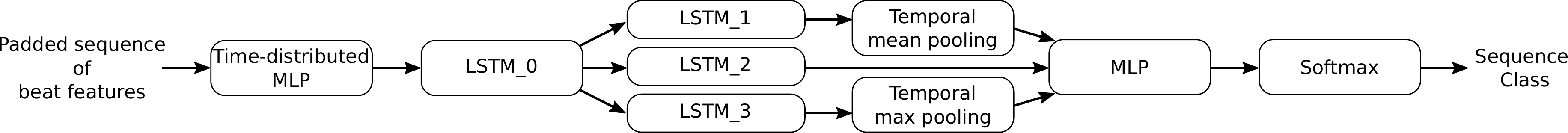}
\caption{The sequence classifier based on LSTMs.}
\label{fig:nnet}
\end{figure*}

The architecture of the sequence classifier is shown in Figure~\ref{fig:nnet}. 
At each time step the features associated with a single heartbeat are 
transformed using a time-distributed multilayer perceptron (MLP). The purpose of 
the MLP is to find a transformation of the input features into a new embedding 
space with easier temporal dynamics. The MLP consists of 256 hidden units 
followed by a Rectifier Linear Unit (ReLU)~\cite{glorot2011deep} which maps the 
input features into a 128 dimensional output space. The \textit{LSTM\_0} network 
further processes the sequence of transformed heartbeat features and outputs 
another sequence that is used as input for the remaining LSTMs. The 
\textit{LSTM\_2} only returns the final state of the network, which is 
subsequently used to output the final classification label. Therefore, the 
central processing pipe of Figure~\ref{fig:nnet} may be seen as a stacked LSTM 
model, which is widely used in machine learning tasks. The processing pipes that 
involve the \textit{LSTM\_1} and \textit{LSTM\_3} networks require further 
explanation. Both LSTMs return new sequences that are transformed into feature 
vectors by using a pooling  operation that drops its temporal dimension. The 
mean pooling averages the outputs of the \textit{LSTM\_1} across its temporal 
dimension, whereas that the max pooling picks the maximum values of each 
dimension of the \textit{LSTM\_3} outputs. Intuitively, what we are trying to 
achieve with these operations is to simulate the reasoning of clinicians that, 
after seeing a complete record, look for some extreme events (the max pooling 
operation) or for some subtle event that occurs during the whole record (mean 
pooling). All the LSTMs used a hidden state of 128 units. Finally, another MLP 
with 256 hidden units and ReLU activation concatenates and processes the outputs 
of the LSTMs before a Softmax layer, which predicts a probability for each of 
the 4 classes. 

The proposed model has a large amount of degrees of freedom and therefore, 
preventing overfitting was a major concern. $L^2$-regularization was applied to 
all layers of the model, using a regularization strength of $10^{-4}$. Dropout 
was also applied to improve generalization. This is achieved by randomly turning 
off several neural units during training, which prevents feature co-adaptation. 
That is, each neural unit becomes more robust since during training time it 
cannot rely on other units to correct its mistakes~\cite{srivastava2014dropout}. 
The amount of dropout was different for each layer of the model: The MLP layers 
used 0.25, \textit{LSTM\_0} used 0.22 for its inputs and 0.44 for its recurrent 
states, and the remaining LSTMs used 0.35 for its inputs and 0.36 for its 
recurrent states. Early stopping was also employed, ending the training after 15 
epochs without improvement on a validation set obtained from a stratified random 
85:15 split. The use of early stopping implies that a small fraction of the data 
would not be used for training the neural network. To use all the available 
data, 3 sequence classifiers were trained using 3 different validation sets. We 
limited the number of LSTMs to 3 to avoid exceeding the Challenge entry size 
limit. The final prediction of the sequence classifier is obtained by averaging 
the predictions of each of the 3 classifiers. Besides using all the available 
data, averaging different models helps in reducing the variance that results 
from the random initialization of the neural network layers and the random split 
between training and validation set. Averaging similar models is also known to 
reduce overfitting. In this sense, this averaging may be seen as a very simple 
sub-bagging.

The sequence classifier was trained by using the categorical cross-entropy as 
the loss function, a batch size of 32 records, and Adam as 
optimizer~\cite{kingma2014adam}. The initial learning rate of the optimizer was 
set to $0.002$, The use of a train and a validation set also permitted us to 
monitor the performance of the classifier and to decrease the learning rate when 
the validation loss was stacked in a plateau. The learning rate was decreased by 
a factor of $\sqrt{2}$ after 3 epochs without improvement.

All the hyperparameters previously discussed were selected by using a tree of 
Parzen estimators, an hyperparameter optimization algorithm based on
approximating the performance of the hyperparameters by using Bayesian 
modeling~\cite{bergstra2013making}. The sequence classifier was implemented 
using Keras~\cite{chollet2015keras}.

\subsection{Classifier stacking}
The global and sequence classifiers were blended together by means of the 
stacking technique. Stacking (or stacked generalization) is an ensemble learning 
method that combines several classifiers by using another meta-classifier 
\cite{wolpert1992stacked}. Stacked classifiers usually achieve better 
performance than the base classifiers by using each level 0 model where it 
performs best. Therefore, the level 0 classifiers should be diverse enough so 
that they may complement each other when combined. Furthermore, the more each 
classifier has to say about the data, the better the resultant stacked 
classifier \cite{wolpert1992stacked}. In our approach, both requirements are 
fulfilled. A Linear Discriminant Analysis (LDA) was chosen as meta-classifier. 
The inputs of the LDA are the predicted probabilities from each of the level 0 
classifiers. However, to avoid possible collinearity issues, only 3 
probabilities from each model are used. An important point to keep in mind when 
training stacked classifiers is that it requires a partition of the dataset. If 
we fitted the level 0 classifiers using all data, then the second 
meta-classifier will be biased towards the best of the two models. Algorithm  
\ref{alg:partitions} describes how we implemented the partitions required by 
both the sequence classifier and the stacking classifier.

\begin{algorithm}
\footnotesize
\caption{algorithm for the stacking classifier. In our implementation we use 
\textit{n\_nets}=3.}
\label{alg:partitions}
\begin{algorithmic}
\INPUT{The \textit{data} consisting on the \textit{Construe} features and the 
classification labels, and the number of neural nets to train \textit{n\_nets}}
\Function{TrainLstms}{train\_data}
    \State{splits = \Call{StratifiedSplit}{train\_data, n\_splits=n\_nets, size=0.15}}
    \State{lstms =  \{\Call{TrainLstm}{train, validation} $\mid$ train, validation $\in$ splits\}}
    \State{\Return{lstms}}
\EndFunction
\State{metafeatures = $\emptyset$\Comment{Probability predictions of each level 0 classifier}}
\For{train\_validation, test  \textbf{in} \Call{StratifiedKFold}{data, n\_folds=6}}
  \State{lstms = \Call{TrainLstms}{train\_validation}}
  \State{lstm\_probs = \Call{Mean}{\{\Call{Predict}{lstm, test} $\mid$ lstm $\in$ lstms\}}}
  \State{xgb = \Call{TrainXgb}{train\_validation}}
  \State{xgb\_probs = \Call{Predict}{xgb, test}}
  \State{metafeatures = metafeatures $\cup$ \Call{Concatenate}{lstm\_probs, xgb\_probs}}
\EndFor
\State{stacking\_classifier = \Call{TrainLda}{metafeatures}}
\State{xgb = \Call{TrainXgb}{data}\Comment{Train the level 0 classifiers with all data}}
\State{lstms = \Call{TrainLstms}{data}}
\OUTPUT{xgb, lstms, stacking\_classifier}
\end{algorithmic}
\end{algorithm}

\section{Results}
\label{sec:results}

Tables~\ref{tab:folds} and~\ref{tab:f1_by_class} show the validation results of 
the classification algorithms in the public and hidden Challenge datasets. 
Table~\ref{tab:folds} shows how the stacking technique usually improves the 
performance of the individual classifiers, achieving a better mean score with 
lower variance. Table~\ref{tab:f1_by_class} shows the same results but 
disaggregated for each target class, where we can see that the magnitude of the 
improvement is almost equal for all classes. Also, Table~\ref{tab:f1_by_class} 
shows the improvement obtained in the follow-up stage of the Challenge with 
respect to the official phase, confirming the change of criteria introduced in 
the bootstrap relabeling and the importance of label consistency. Yet, the 
notable performance decrease in the hidden test set with respect to 
cross-validation suggest that there are still criteria differences between the 
training and test sets.

\begin{table}[!ht]
\vspace{-1em}
\footnotesize
\caption{Example of stratified 8-fold cross-validation}
\label{tab:folds}
\begin{center}
\begin{tabular}{ c r r r r r r r r c}
\hline \hline
 & \multicolumn{8}{c}{Fold Number}& \multirow{2}{*}{Mean (SD)}\\
\cline{2-9}
Method & 0&  1& 2& 3& 4& 5& 6& 7&\\
\hline 
XGBoost & 0.889 & 0.874 & 0.862 & 0.866 & 0.871 & 0.905 & 0.908 & 0.867 
& 0.880 (0.018)\\
LSTMs & 0.866 & 0.868 & 0.849 & 0.862 & 0.848 & 0.870 & 0.886 & 0.862 
& 0.864 (0.012)\\
LDA-stacker & 0.904 & 0.883 & 0.872 & 0.887 & 0.872 & 0.901 & 0.905 & 0.886 
& 0.889 (0.013) \\
\hline \hline
\end{tabular}
\end{center}
\vspace{-2em}
\end{table}

\begin{table}[!ht]
\caption{Per-class performance in training and test sets}
\label{tab:f1_by_class}
\footnotesize
\begin{center}
\begin{tabular}{|c|r|r|r||r|r|}\hline
   & \multicolumn{3}{c||}{Training set cross-validation mean $F_1$ (SD)} & 
     \multicolumn{2}{c|}{Test set $F_1$}  \\\hline
   & \multicolumn{1}{c|}{\multirow{2}{*}{XGBoost}} & 
     \multicolumn{1}{c|}{\multirow{2}{*}{LSTMs}} & 
     \multicolumn{1}{c||}{\multirow{2}{*}{LDA-stacker}} & 
     \multicolumn{2}{c|}{LDA-stacker}\\\cline{5-6}
   & & & & Official phase & Follow-up \\\hline
\textbf{N} & 0.953 (0.007) & 0.955 (0.006) & 0.960 (0.007) & 0.90 & 0.92\\\hline
\textbf{A} & 0.838 (0.032) & 0.796 (0.028) & 0.842 (0.021) & 0.85 & 0.86\\\hline
\textbf{O} & 0.850 (0.019) & 0.841 (0.012) & 0.864 (0.019) & 0.74 & 0.77\\\hline
$\sim$     & 0.711 (0.063) & 0.658 (0.052) & 0.724 (0.045) & - & -\\\hline
\textbf{Overall} & \textbf{0.880 (0.018)} & \textbf{0.864 (0.012)} & 
\textbf{0.889 (0.013)} & \textbf{0.83} & \textbf{0.85}\\\hline
\end{tabular}
\vspace{-1.5em}
\end{center}
\end{table}

\section{Discussion and conclusions}
\label{sec:conclusions}

The inconsistency problems in the Challenge dataset suggest that giving complete 
independence to human experts leads to a high level of disagreement in the 
resulting labels, even in a relatively bounded and well-defined problem from the 
clinical perspective like the ECG arrhythmia classification. However, the 
presence/absence of diseases like atrial fibrillation is not a subjective 
consideration, and medical guidelines prove that consensuses can be 
achieved~\cite{Camm10}. We defend that such a consensus should be the first step 
for the development of an effective low-cost screening system, as it removes the 
need for dozens of experts and thousands of records to achieve an statistically 
acceptable agreement, while allowing for less complex algorithms. We would like 
to point out that by ``consensus'' we are not referring to a strong 
quantification of decision thresholds, but to general guidelines shared by the 
expert labelers. A possible example could be: \textit{``Atrial flutters will be 
considered in the \textbf{O} class''}, which reduces the probability of 
inconsistencies such as the one in Figure~\ref{subfig:flutter}.

Having these guidelines, and on the basis of the achieved results in the 
Challenge, we consider that a reduced set of two or three experts and some 
hundreds of records should be enough to achieve a much higher performance of the 
classification algorithm, with an $F_1$ over 0.9. Additionally, it would be 
desirable that labeling would be made on standard multi-lead ECG records 
acquired simultaneously to the single-lead record, to quantify the information 
loss of the handheld device for this specific problem. We expect to conduct this 
study in the near future.

The interpretation of ECG signals is a task in which humans show an outstanding 
proficiency. But even if qualities like personal experience or intuition are 
invaluable and hardly formalizable, the pathophysiological processes that can be 
observed in the ECG and the effects they have on the signal behavior are widely 
accepted in the medical community. Also, systematic approaches such as those 
described in ECG handbooks~\cite{Marriott08} provide excellent results for 
screening and diagnosis, and most of that expert knowledge is susceptible of 
being formalized by computational methods.

The official results of the Challenge show that all best-performing algorithms 
include domain-specific knowledge at some point. This suggest that 
knowledge-based approaches have a fundamental advantage over pure learning-based 
approaches in quantifying the underlying criteria of manually labeled datasets. 
In the present proposal, we demonstrated that exploiting this advantage is 
feasible without sacrificing the benefits of sophisticated machine learning 
methods and maintaining a notable degree of interpretability by the use of 
meaningful features.

Finally, in this Challenge the \textit{Construe} algorithm has reaffirmed its 
ability to accurately interpret highly contaminated data by combining bottom-up 
reasoning (guided by data) and top-down reasoning (guided by knowledge) in an 
abductive cycle. As forthcoming work, the integration of new signal types beyond 
the ECG, such as for example the blood pressure, has the potential to further 
improve the robustness and the accuracy of the results.


\section*{Acknowledgements}  
This work was supported by the Spanish Ministry of Economy and Competitiveness 
under project TIN2014-55183-R. C.A. Garc\'ia is also supported by the 
FPU Grant program from the Spanish Ministry of Education (MEC) (Ref. 
FPU14/02489). We would also like to acknowledge Alberto Cobos for his work on 
the lead inversion detection problem during his CiTIUS summer fellowship 2017, 
and the Challenge organizing team for the great support and encouragement given 
to the competitors at all times.

\section*{References}
\bibliographystyle{dcu}
\bibliography{bibliography}

\end{document}